\documentclass{article}

\usepackage{arxiv}

\usepackage[utf8]{inputenc} 
\usepackage[T1]{fontenc}    
\usepackage{hyperref}       
\usepackage{url}            
\usepackage{booktabs}       
\usepackage{amsfonts}       
\usepackage{nicefrac}       
\usepackage{microtype}      
\usepackage{lipsum}
\usepackage{graphicx}
\usepackage{amsmath,amssymb,amsfonts}

\usepackage{enumitem}
\usepackage{algorithm}
\usepackage{graphicx}
\usepackage{subfig}
\usepackage[noend]{algpseudocode}

\title{Learning Social Heuristics for Human-Aware Path Planning}

\author{
 Andrea Eirale \\
  Department of Electronics and Telecommunications (DET)\\
  Politecnico di Torino\\
  Torino, Italy\\
  \texttt{andrea.eirale@polito.it} \\
   \And
 Matteo Leonetti \\
  Department of Informatics\\
  King’s College London\\
  London, UK\\
  \texttt{matteo.leonetti@kcl.ac.uk} \\
  \And
 Marcello Chiaberge \\
  Department of Electronics and Telecommunications (DET)\\
  Politecnico di Torino\\
  Torino, Italy\\
  \texttt{marcello.chiaberge@polito.it} \\
}

\begin{document}
\maketitle
\begin{abstract}

Social robotic navigation has been at the center of numerous studies in recent years. Most of the research has focused on driving the robotic agent along obstacle-free trajectories, respecting social distances from humans, and predicting their movements to optimize navigation.
However, in order to really be socially accepted, the robots must be able to attain certain social norms that cannot arise from conventional navigation, but require a dedicated learning process.
We propose Heuristic Planning with Learned Social Value (HPLSV), a method to learn a value function encapsulating the cost of social navigation, and use it as an additional heuristic in heuristic-search path planning. In this preliminary work, we apply the methodology to the common social scenario of joining a queue of people, with the intention of generalizing to further human activities.

\end{abstract}


\section{Introduction} \label{sec:intro}
In the last few years, the proliferation of robotic platforms in society has inspired a significant amount of research in the field of service robotics. Robots can provide support to humans in a large number of specific tasks, such as helping the elderly \cite{martinez2018personal, vercelli2018robots, eirale2022marvin}, giving classes to students \cite{lee2011effectiveness, toh2016review}, giving guided tours \cite{shiomi2006interactive, al2016tour, sasaki2017long} and many more \cite{pieska2013social, kyrarini2021survey}. 

In all these cases, how the robot interacts with people and how human beings perceive it is often fundamental for the successful completion of the service task \cite{honour2021perceived}. Thus, traditional autonomous navigation, focused on avoiding obstacles while reaching the goal, is often unsuitable. In addition to conventional navigation metrics, like the distance to the goal or the presence of static obstacles, robots can exploit navigation behaviors that consider social factors. Introducing social information can lead to navigation policies able to access a trade-off between performance and social acceptability. The robot should not only respect others' personal spaces, but also comply with specific social norms. These norms include planning aspects, such as navigating through cluttered and crowded environments, and behavioural aspects, namely the adaptation to social signals which can be inferred from human actions \cite{mavrogiannis2023core}.

We propose a novel method to augment classical path planning with socially acceptable behaviours learned through Reinforcement Learning (RL), at no additional planning cost. 
We train a social agent able to recognize specific social contexts; as a proof of concept, for this paper, we consider the problem of following a queue. The learned value function encapsulates the complexity of the social behaviour and depends on all necessary external factors, such as the position and activity of other people in the environment, so that the path planner does not have to take them into account. The learned value function is combined with the heuristic used by A*, which, with no change to the planning algorithm, produces socially acceptable trajectories.

\section{Related Work} \label{sec:related}
The problem of designing a behavioral policy able to consider social factors has been studied for more than twenty years. Following the RHINO \cite{burgard1999museum} and MINERVA \cite{thrun2000probabilistic} deployments as tour guides in museums, where people are treated like dynamic, non-responsive obstacles, researchers focused on autonomous navigation systems able to distinguish humans from inanimate objects.

An important paradigm in this area involves planning around estimates of future human motion, which allows the agent to plan an optimal trajectory avoiding collisions with people and obstacles, and maintaining at any time a social distance from humans. Human motion prediction is usually achieved with deep neural networks, like generative adversarial networks \cite{gupta2018social}, convolutional neural networks \cite{mohamed2020social, zhao2020noticing} and attention transformers \cite{vemula2018social}.

Another central body of work, which extends human motion prediction, consists in intention-aware navigation. Exploiting behaviour prediction models, it is possible not only to predict future movements of humans, but also to estimate their final goals. Bai er al. \cite{bai2015intention} model the uncertainty of human intent in the Partially Observable Markov Decision Process (POMDP) framework. Other works focused on navigation in dense human crowds avoiding blockage due to human activities \cite{park2016hi}, and cooperating with people through interacting Gaussian processes \cite{trautman2015robot}. Mavrogiannis et al. employ geometric \cite{mavrogiannis2019multi} and topological invariant \cite{mavrogiannis2020multi} representations to model the coupling among trajectories of multiple navigating agents.

A different class of works have proposed Deep Reinforcement Learning for prediction in crowd navigation domains. Chen et al. \cite{chen2017decentralized} apply CADRL, a deep reinforcement learning framework for socially aware multiagent collision avoidance. Everett et al. \cite{everett2018motion} exploit an actor-critic variant to relax prior assumptions and learn policies and agent motion models at the same time. Furthermore, Tai et al. \cite{tai2018socially} train a generative adversarial imitation learning model on a dataset generated using the social force model \cite{helbing1995social}. Finally, Chen et al. \cite{chen2019crowd} use attention-based reinforcement learning to produce interaction-aware collision avoidance behaviors.

All the work cited above focuses on people movement: their trajectory, their intention and their destination. However, a number of social norms are independent of movement, but dependent on people's activity. For instance, it is considered rude to walk between two people talking to each other, even if they are just standing still. Our work is an initial step in the direction of integrating arbitrary scene features (not necessarily movement) into classical path planning.

Within queue following, the particular social navigation scenario we consider in this paper, Nakauchi et al. \cite{nakauchi2002social} designed a dedicated pipeline. In their work, they put great attention on how a line of people can be defined and then perceived by the autonomous agent. The navigation system generates a series of goals (depending on the number of people in the queue) until the back of the line is reached, and does not provide full path planning.

\section{Methodology} 
The main insight of our approach is that the social component, which depends on people in the environment and their activities, and the navigation component, which only depends on position and orientation of the robot, can be decoupled. The social component is trained off-line with RL, and then re-integrated in the planner's objective function. We start by introducing the notation regarding task modeling and heuristic-search planning.

\label{sec:methodology}

\subsection{Notation} \label{subsec:notation}
We model the navigation task as a Markov Decision Process (MDP) described by the tuple $(\mathcal{S},\mathcal{A}, \mathcal{P}, R, \gamma)$ \cite{sutton2018reinforcement}. An agent starts its interaction with the environment in an initial state $s_0$, drawn from a distribution $p(s_0)$ and then at every time step $t$ selects an action $a_t \in \mathcal{A}$ from a state $s_t \in \mathcal{S}$ which results into a new state $s_{t+1}$, receiving a reward $r_t = R(s_t,a_t)$. A reinforcement learning process aims to optimize a parametric policy $\pi_\theta$, which defines the agent behavior. In the context of navigation learning, we model the task as an episodic MDP, with maximum time steps $T$. Hence, the agent is trained to maximize the cumulative expected reward $\mathbb{E}_{\tau\sim\pi} \sum_{t=0}^{T} \gamma^t r_t$ over each episode, where $\gamma \in [0,1)$ is the discount factor. 

We use heuristic-search planners~\cite{ferguson2005guide}, such as A*, which estimate the value of a given state $s$ as the sum of the cumulative cost up to $s$ and a heuristic value from $s$, minimizing the objective: 
\begin{equation}
\label{planner_objective}
f(s) = g_n(s) + h_n(s),
\end{equation}
where $n$ indicates that both functions are related to the navigation objective.
We will modify this objective by adding a social component.

\begin{figure*}[t]
    \centering
    
    \subfloat[][\emph{Traditional $A^*$}]
        {\includegraphics[width=52mm]{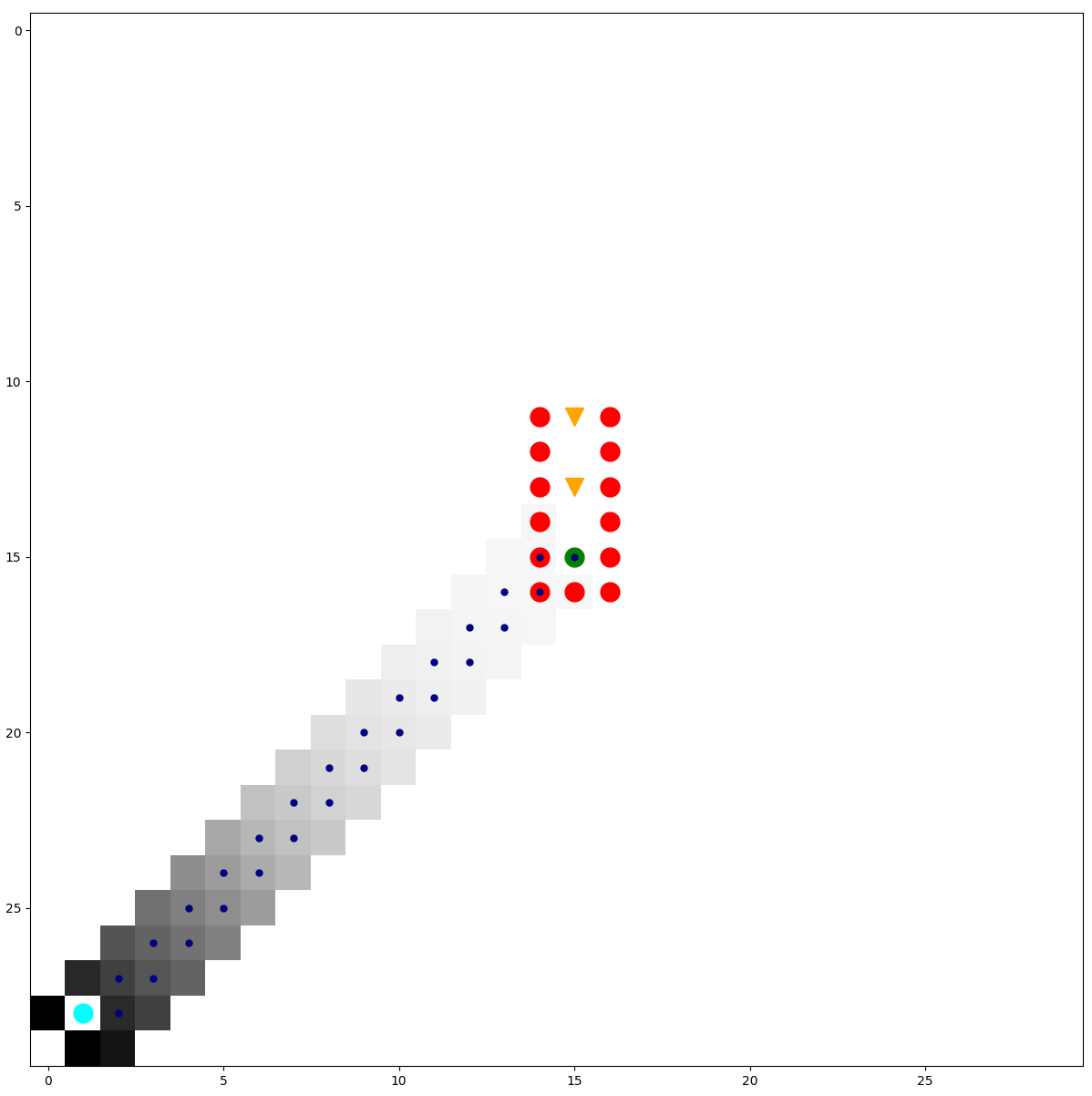}} \quad
    \subfloat[][\emph{Social Cost Function}]
        {\includegraphics[width=52mm]{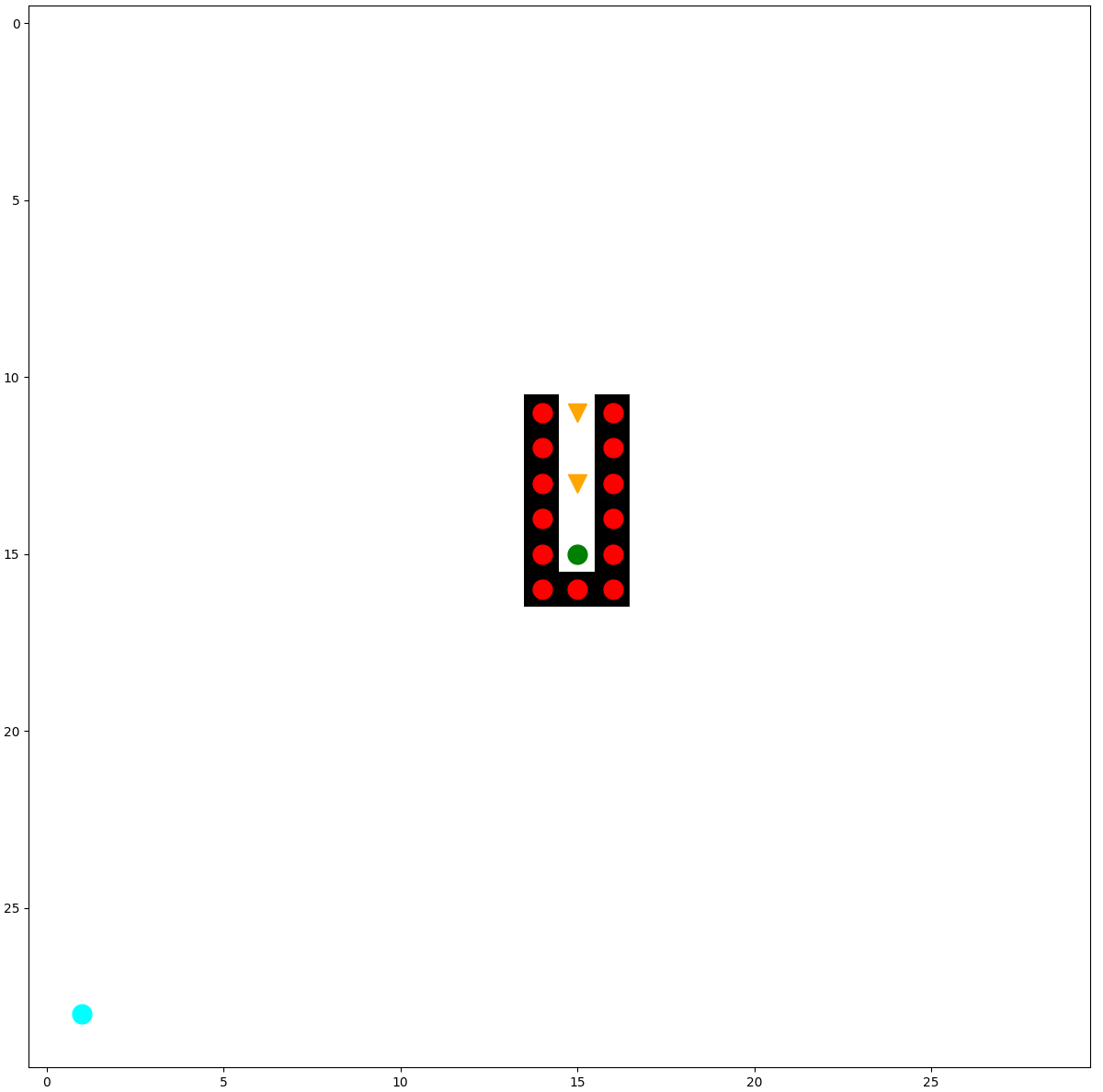}} \quad
    \subfloat[][\emph{Integration of $A^*$ with Social Cost Function}]
        {\includegraphics[width=52mm]{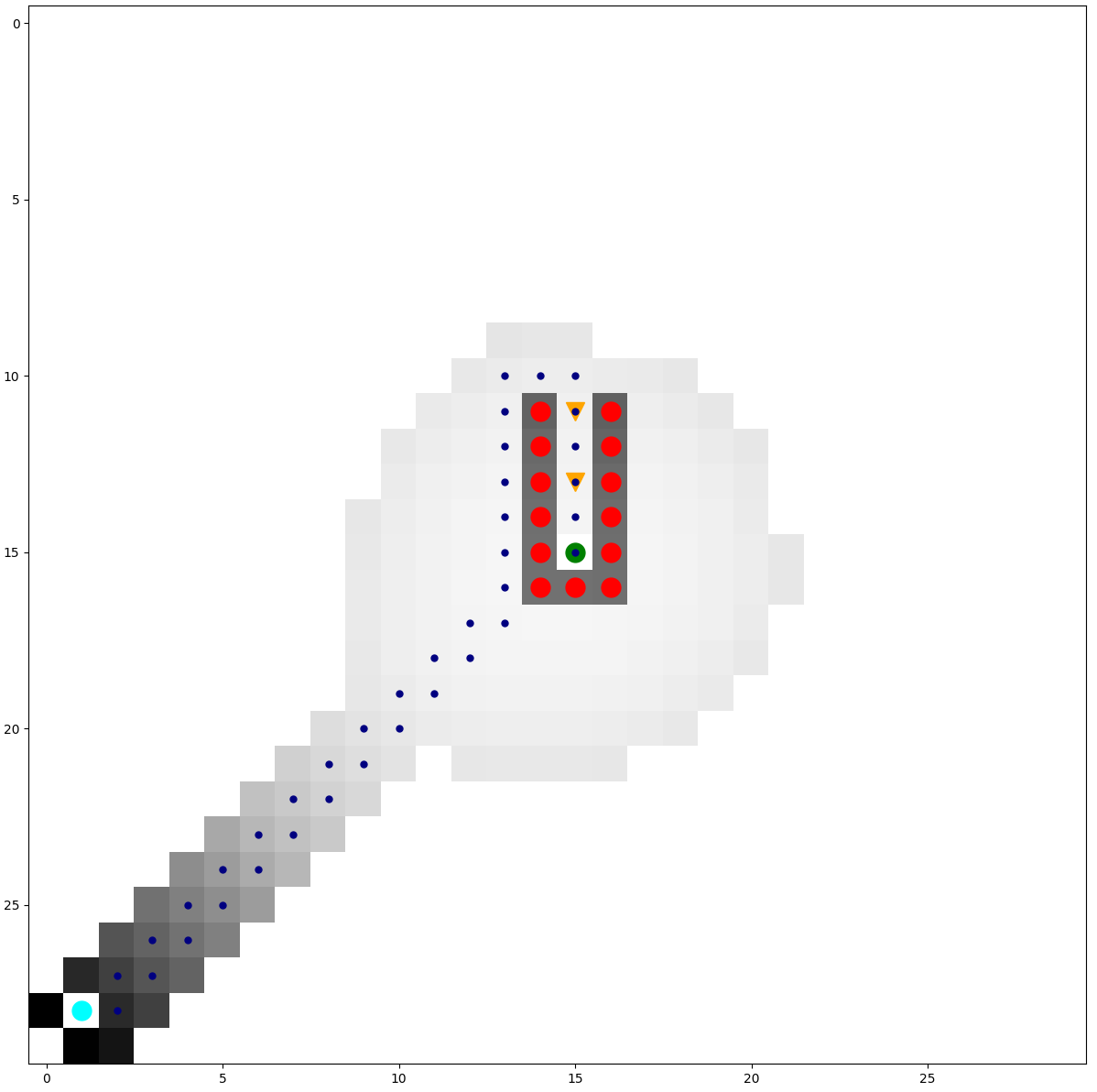}} \\
        
    \caption{Results obtained from the demo discrete environment: cyan and green points are respectively the start and goal points, orange arrows represent people, small blue dots are the trajectory chosen by the planner, and red dots are the virtual obstacles. Black cells represent cells associated with a high cost, while the cost decrease as they turn to light grey. $(a)$ represents the path chosen by a traditional $A^*$ search algorithm, and (b) shows the costs added by the Social Cost Function $c_s$ extracted from $Q_s$. In contrast, (c) represents the path chosen by the $A^*$ algorithm integrated with the Social Cost Function $c_s$.}
    \label{fig:discrete_results}
\end{figure*}

\subsection{Planning Objective}

In addition to the usual navigation cost and heuristic (cf. Eq. \ref{planner_objective}), we want the planner to take also a \emph{social} cost into account when computing the optimal path. A social cost would be non-zero, for instance, when a position would result in the robot cutting a queue, or driving between two people in a conversation. In this work, we assume the existence of such a social cost function $c_s(s)$, which, in general, is also a function of the goal. In principle, it is also possible to learn this cost function from experience, but for this preliminary work we assume is as known.

The social cost makes the problem effectively multi-objective: the robot has to both minimize travel distance and social cost. Since heuristic-search planners require a heuristic function in addition to the cost function, and the social cost is known, the rest of this methodology is devoted to the definition of a social heuristic.

We modify the objective function of Equation \ref{planner_objective} to take both objectives into account, representing with the subscript $n$ the navigation component of the objective and with the subscript $s$ the social component of the objective:
\begin{equation}
\label{planner_modified_objective}
f(s) = g_n(s) + h_n(s) + w(g_s(s) + h_s(s)),
\end{equation}
where $w \geq 0$ is a parameter weighting the social cost over the navigation cost.

The function $g_s$ is obtained by accumulation during planning of the cost $c_s$, just like for its navigation counterpart. The function $h_s$ is defined in the following section.

\subsection{Social Heuristic} 
\label{subsec:training}
We propose to obtain the social heuristic function $h_s$ through RL.
\begin{description}[leftmargin=0pt]
\item[Environment] The agent is trained in a fully observable gridmap-like environment, with a given number of people whose position and orientation in the grid is known.

\item[State Space] The state representation embeds the necessary information about the goal and people, and it is ego-centric, so that it does not depend on the particular coordinates used in training. Different definitions are possible, depending on the social behaviour one intends to capture. For our queue following scenario we used:
\begin{itemize}
    \item The distance $d_t^g$ and the angle $\Delta\theta_t^g$ of the goal from the agent.
    \item For every person in the environment, the distance $d_t^i$ and the angle $\Delta\theta_t^i$ of the $i$th person with respect to the agent.
\end{itemize}
The state space is, therefore, defined as:
\begin{equation}
\label{state_space}
    S(t) = [d_t^g, \Delta\theta_t^g, d_t^1, \Delta\theta_t^1, ..., d_t^i, \Delta\theta_t^i].
\end{equation}
\item[Action Space] The agent can choose to go forward, backward, turn left, or right. 
\item[Reward]  We define the navigation reward $r_n$ as:
\begin{equation}
    r_n = \begin{cases}
    \multicolumn{1}{@{}c@{\quad}}{k_g} & \text{if goal is reached} \\
    d_{t-1}^g - d_t^g - k_r & \multicolumn{1}{@{}c@{\quad}}{\text{otherwise,}}
    \end{cases}
\end{equation}
where $d_{t}^g$ is the distance of the goal from the agent, $d_{t-1}^g$ is the distance of the goal at the previous time step, $k_g$ is a large reward received when the goal is reached, and $k_r$ is a small value implemented to encourage convergence in as few steps as possible. 
The social reward is the opposite of the weighted social cost: $r_s = -w c_s$.

We train an RL agent to maximize $r_T = r_n + r_s$ with standard RL, and, in doing so, it learns a value function $Q_T$. At the same time, the agent learns a second value function $Q_s$, trained to estimate the expected cumulative unweighted social component of the reward. This social value function is trained on reward $r_s$ only. 
Note that, during learning, actions are only selected so as to maximize $Q_T$. That is, the function $Q_s$ is learned entirely as a side effect of maximizing $Q_T$, and does not affect the behaviour policy. This is crucial because $Q_s$ does not encode the navigation goal. For instance, in following a queue, maximizing $Q_T$ leads the agent to wait at the end of the queue, while maximizing $Q_s$ would only lead the agent to stay away from people in order to avoid the potential cost of cutting the queue. So, while executing and learning the full task, the agent also stores a long-term estimate of the social cost for later use as a planning heuristic.

\end{description}

\subsection{Deployment} \label{subsec:deployment}

In the deployment phase, the value function $Q_T$ is discarded, since we are only interested in the social components of the objective function $g_s$ and $h_s$ as in Equation \ref{planner_modified_objective}. These components are extracted from the social value function $Q_s$. In particular, the heuristic social component is computed as:

\begin{equation}
   h_s(s) = \begin{cases}
   c_s(s) & \text{if } c_s(s) > k_{thresh} \\
   \multicolumn{1}{@{}c@{\quad}}{0} & \multicolumn{1}{@{}c@{\quad}}{\text{otherwise}}
   \end{cases}
\end{equation}
\begin{equation}
   c_s(s) = 1 - Q_s(s, a_{s^-\rightarrow s}, \theta_s)
\end{equation}

Where $s_t$ is the agent observation at time $t$, $a_{s^-\rightarrow s}$ is the action that brings the agent from the current state $s^-$ to the next state $s$, $\theta_s$ are the trained social wights, and $Q_s(s_t, a_{s^-\rightarrow s},\theta_s)$ is the value predicted by the action-value function $Q_s$ when the action $a_{s^-\rightarrow s}$ is taken starting from the state $s^-$. Finally, $k_{thresh}$ is a confidence threshold: if the cost $c_s(s)$ exceeds this value, the agent is making a strong assumption on which direction (not) to take. This configuration allows the $A^*$ algorithm to manage the general path planning, while the social component adds its contribution only when the social scenario is recognized.

Moreover, the cumulative social cost $g_s(s)$ is defined as:
\begin{equation}
   g_s(s) = g_s(s^-) + c_s(s)
\end{equation}\newline





\section{Experiments and Results} \label{sec:results}
To validate our methodology, we trained a HPLSV agent exploiting a customized version of the TF2RL library \cite{ota2020tf2rl}. As a proof of concept, we consider the social scenario of a queue of people, where the agent has to enter the queue without cutting it. At first, the agent is trained in a demo gridmap environment, where people are perfectly aligned with the goal, and the distance between the goal and the first person, and between each person, has the same value. Then, a series of close-to-real environments are obtained from a Gazebo simulation, and are used to retrain the agent. In the next paragraphs, the results obtained in each of these environments are presented.

\subsection{Discrete environment} \label{subsec:demo}
The first environment presents a 30x30 gridmap. The goal is placed in the center of this gridmap, and two other entities, representing people in line, are placed just above the goal. One cell separates the goal from the first person, and another cell is left between the two people. The three entities, the goal and the two people, are aligned on the x axis. We define virtual obstacles as a continuous box surrounding people and goal, which leaves just a passage to the goal at the end of the queue. If the agent hits the virtual obstacle, it is equivalent to cutting the queue, and it receives the corresponding social cost. During initial training episodes, the agent always starts from the same position, at the bottom left of the gridmap. Then, in order to enhance exploration, the starting point is changed after each episode, until the end of the training. For the considered scenario of a queue of people, the social reward function $r_s$ is defined as:

\begin{equation}
   r_s = \begin{cases}
   \multicolumn{1}{@{}c@{\quad}}{-k_s} & \text{if the virtual obstacle is crossed} \\
   \multicolumn{1}{@{}c@{\quad}}{0} & \multicolumn{1}{@{}c@{\quad}}{\text{otherwise}}
   \end{cases}
\end{equation}

Where $k_s$ is a small value received every time the agent cross a virtual obstacle.\\

To test and validate the model, it is integrated with a classic $A^*$ planner, as explained at Section \ref{subsec:deployment}. Figure \ref{fig:discrete_results}a shows the path chosen by the $A^*$ planner without any other external contribution, with the related costs. In this case the planner just drives the agent towards the goal, cutting the line. Figure \ref{fig:discrete_results}b shows the costs added by the social action-value function $Q_s$ extracted from the trained model. Finally, Figure \ref{fig:discrete_results}c shows the path and the costs obtained by the $A^*$ planner integrated with our trained model. In contrast with the previous case, the planner is now able to recognize the social scenario and extract an optimal path to the goal which passes through the end of the queue, avoiding to cut it.

\begin{figure*}[t]
    \centering
    
    \subfloat[][\emph{}]
        {\includegraphics[width=0.47\columnwidth]{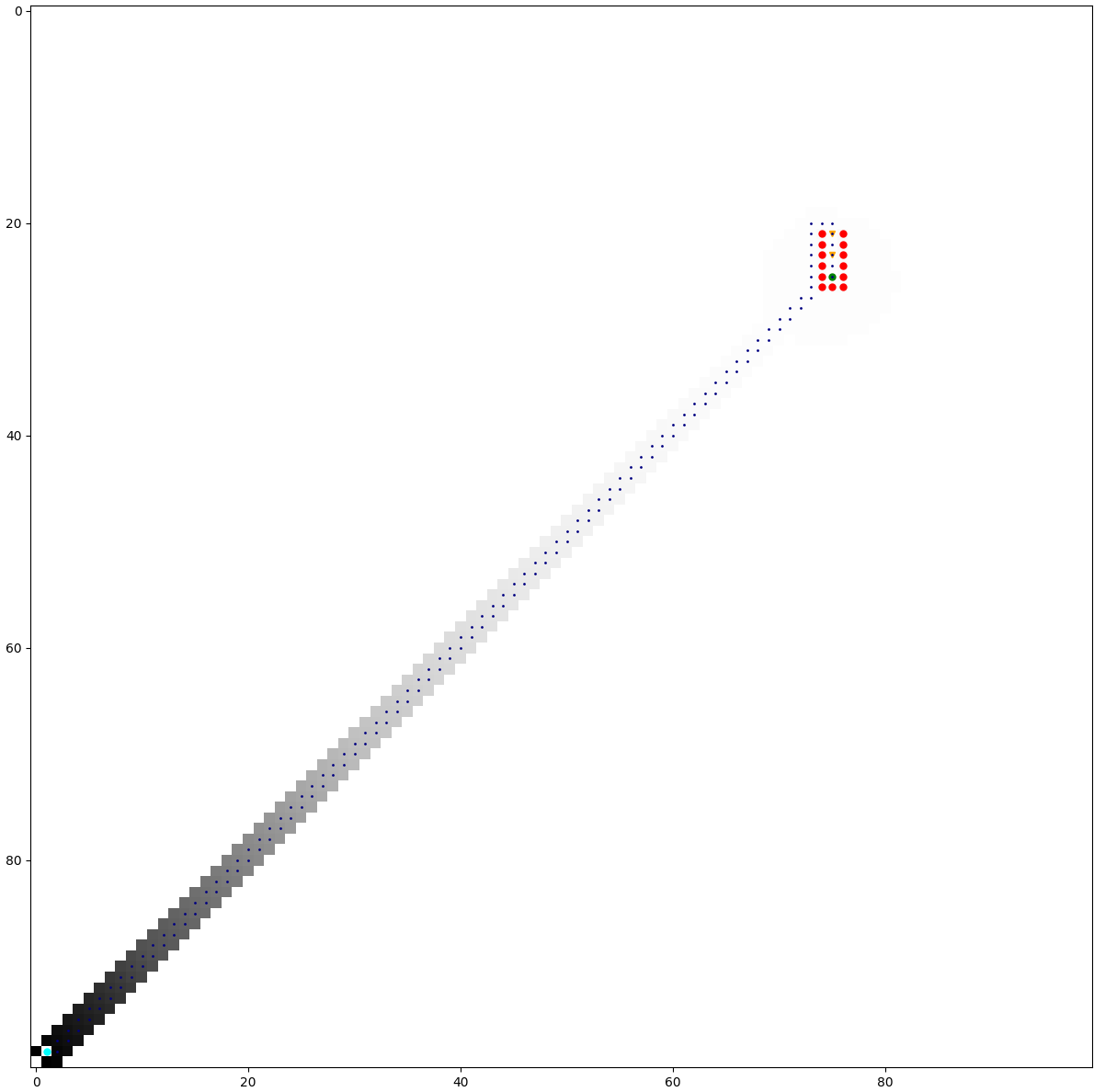}} \quad
    \subfloat[][\emph{}]
        {\includegraphics[width=0.47\columnwidth]{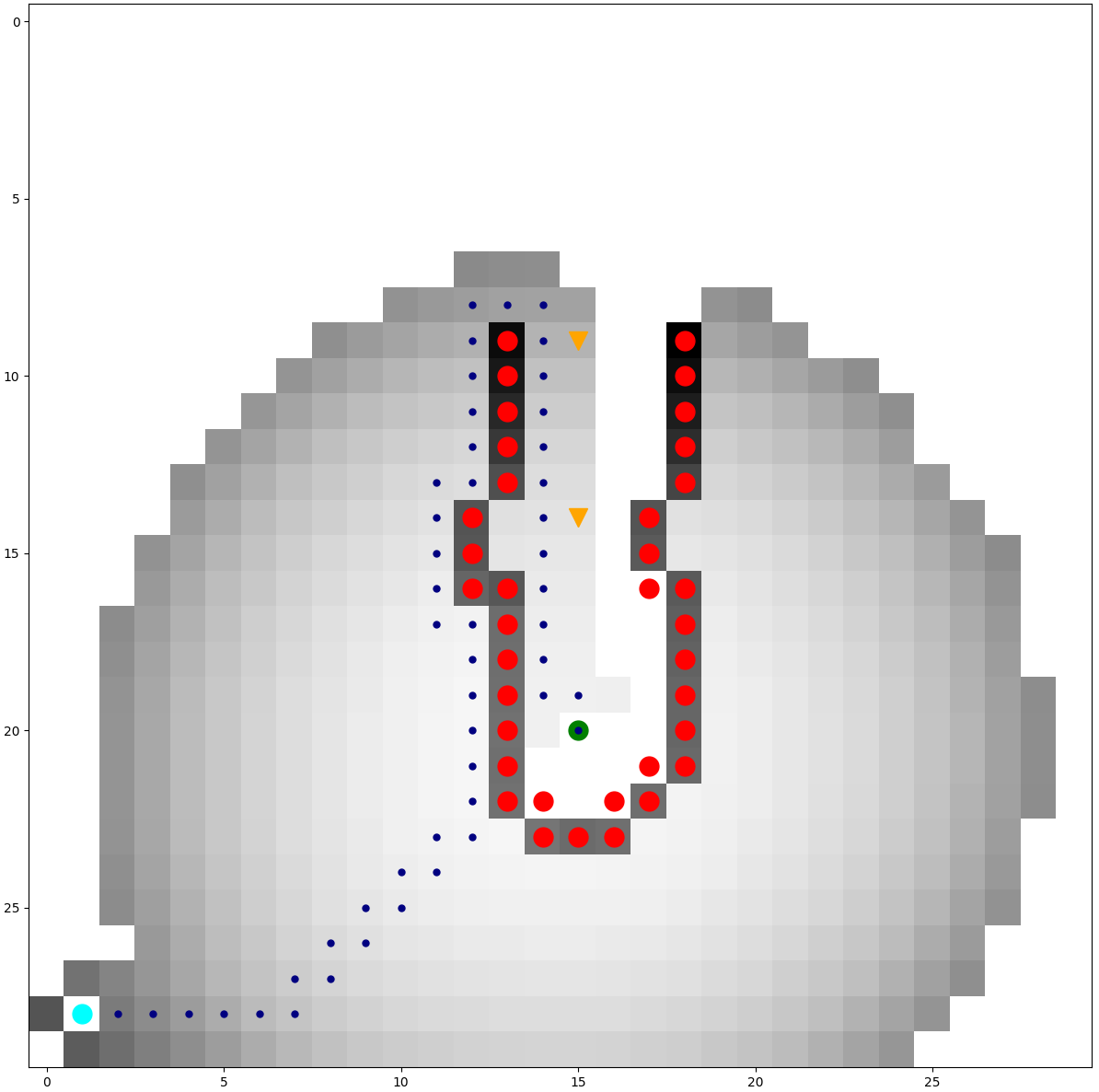}}\\
        
    \caption{In (a), the results obtained on a much bigger environment compared to the one used for training. The social cost function does not interfere with the traditional planner until the social scenario is reached. In (b), Results obtained from the continuous environment.}
    \label{fig:other_results}
\end{figure*}

The ego-centric representation of the heuristic makes it independent of coordiantes, as shown in Figure \ref{fig:other_results}(a), where a much larger 100x100 gridmap testing environment is employed. The social component of the objective does not activate (the expected social cost of actions is $0$) until the agent gets near the goal. This allows the traditional planner to compute an optimal trajectory towards the goal without any other interference. Then, when the queue is reached, the costs introduced by the social contributions increase, avoiding to cut the line.

\subsection{Continuous environments} 
A series of different environments are obtained from a Gazebo simulation, to train and test the agent in more realistic conditions. People and goals are manually set in a continuous environment, in order to recreate a realistic queue scenario. Each of these environments are then discreteized into a 30x30 gridmap with a resolution of 0.2 meters. The virtual obstacles are defined as a box similar to demo version, at a distance of 0.5 meters from the goal and from each person. The new obtained discrete environments are then used to retrain the previous DRL agent. During the training both, the starting point of the agent and the environment, are randomly changed after every episode.
Similarly to before, the model is integrated with the $A^*$ planner, and the results can be observed in Figure \ref{fig:other_results}(b).

\section{Conclusions} \label{sec:conclusions}
We introduced Heuristic Planning with Learned Social Value (HPLSV), a novel RL-based human-aware path planning method. For these preliminary results, we developed a proof of concept on queue following, but intend to extend the methodology to more, if not all, human activities.

The greater limitation of our approach resides in the choice of a proper representation for the working environment. In this work we assumed to know the exact position and orientation of all and only the people in the queue. In the real world, a scene may have many people, most of which have nothing to do with the navigation task. Recognising and interpreting human activity is part of the social navigation challenge, and we do not expect the heuristic function to be able to solve this problem end-to-end, directly for robot sensory inputs. Another strong assumption is the a priori knowledge of the social cost function, represented in training with virtual obstacles. In general, it would be interesting to learn it from real human-robot interactions.

\bibliographystyle{unsrt} 
\bibliography{biblio}

\end{document}